\documentclass{article}

\usepackage[preprint, nonatbib]{nips_2018}

\usepackage{float}

\usepackage[square,numbers]{natbib}
\usepackage[utf8]{inputenc} 
\usepackage[T1]{fontenc}    
\usepackage{hyperref}       
\usepackage{url}            
\usepackage{booktabs}       
\usepackage{amsfonts}       
\usepackage{nicefrac}       
\usepackage{microtype}      
\usepackage{graphicx}
\usepackage{tabularx}
\usepackage{hyperref}

\usepackage{amsmath}
\usepackage{algorithm}  
\usepackage{algpseudocode} 
\usepackage{subfigure}
\usepackage{multirow}

\title{The Neural Metric Factorization for Computational Drug Repositioning}

\author{
	Xinxing Yang \\
	
	Ningbo Artificial Intelligence Institute, Shanghai Jiao Tong University\\
	Department of Automation, Shanghai Jiao Tong University\\
	\texttt{yangxinxing@sjtu.edu.cn} \\
	
	\And
	Genke Yang \\
	Ningbo Artificial Intelligence Institute, Shanghai Jiao Tong University\\
	Department of Automation, Shanghai Jiao Tong University\\
	\texttt{gkyang@sjtu.edu.cn} \\
	
	\And
	Jian Chu \\
	Ningbo Artificial Intelligence Institute, Shanghai Jiao Tong University\\
	Department of Automation, Shanghai Jiao Tong University\\
	\texttt{chujian@sjtu.edu.cn} \\
	
}

\begin{document}
\maketitle

\begin{abstract}
Computational drug repositioning aims to discover new therapeutic diseases for marketed drugs and has the advantages of low cost, short development cycle, and high controllability compared to traditional drug development. The matrix factorization model has become the cornerstone technique for computational drug repositioning due to its ease of implementation and excellent scalability. \par
	
However, the matrix factorization model uses the inner product to represent the association between drugs and diseases, which is lacking in expressive ability. Moreover, the degree of similarity of drugs or diseases could not be implied on their respective latent factor vectors, which is not satisfy the common sense of drug discovery. \par
	
Therefore, a neural metric factorization model (NMF) for computational drug repositioning is proposed in this work. We novelly consider the latent factor vector of drugs and diseases as a point in the high-dimensional coordinate system and propose a generalized Euclidean distance to represent the association between drugs and diseases to compensate for the shortcomings of the inner product. Furthermore, by embedding multiple drug (disease) metrics information into the encoding space of the latent factor vector, the information about the similarity between drugs (diseases) can be reflected in the distance between latent factor vectors. Finally, we conduct wide analysis experiments on two real datasets to demonstrate the effectiveness of the above improvement points and the superiority of the NMF model. \par
	
\end{abstract}


\section{Introduction}

Despite the rapid advances in technologies such as genomics and bioinformatics in recent years, the process of developing new drugs is still time-consuming and costly. According to the work of Dickson et al \cite{1,2}, the average period of new drug development is about 10-15 years and costs about 0.8-1.5 billion dollars. In addition, new drugs have unknown side effects, most of them do not successfully pass clinical trials. To overcome the limitations of traditional new drug development, Ashburn et al. \cite{3} proposed computational drug repositioning techniques for mining potential therapeutic diseases of marketed drugs. This technology has been shown by Thor et al. \cite{3} to be an excellent and affordable strategy for accelerating new drug discovery, reducing the drug development cycle to 6.5 years and the required R\&D funding to 3 million dollars \cite{4,5}. The medical principle that underpins computational drug repositioning is that drugs enter the body to treat diseases by activating or inhibiting certain targets, and that marketed drugs can react with multiple unknown targets, which can lead to the existence of some unknown therapeutic diseases \cite{6}. Therefore, compared with the traditional drug development process, computational drug repositioning technology can significantly accelerate the drug development process, save the investment cost and enhance drug controllability, which have great practical and economic value to the pharmaceutical industry \cite{7}. \par

The purpose of computational drug repositioning is to infer new drug-disease interaction associations based on known drug-disease associations using various predictive models \cite{8,9,10}. Overall, the current mainstream computational drug repositioning models fall into two main categories, one is network-based models \cite{n1,n2,n3,n4}. The other is matrix factorization-based models \cite{m1,m2,m3,m4,mf1,mf2}. \par

The network-based model identifies the set of similar neighbors of target drug or target disease by the historical association information. And then recommends potential therapeutic diseases to the drug based on this set of neighbors. Cheng et al. \cite{n1} proposed a network-based inference model (NBI) for computational drug repositioning, which differs from traditional approaches based on the structure of drug molecules, protein targets, or disease phenotypic features. The NBI model exploits the topological similarity of drug-disease association networks to infer potential new uses of known drugs, providing a new solution for the field of computational drug repositioning. To combine heterogeneous network information with the field of computational drug repositioning, Wang et al. \cite{n2} proposed TL-HGBI, a computational framework based on heterogeneous network model, which first constructs heterogeneous network using drug-disease associations and drug-target relationships. Then, based on the principle of concatenation, an iterative algorithm of the heterogeneous network is used to predict potential drug-disease associations. The novelty of this framework is that it can calculate the strength of drug-disease associations by iterative algorithms of heterogeneous network and incorporates target information in the calculation process. Based on the assumption that similar drugs can treat similar diseases, Luo et al. \cite{n3} proposed the MBiRW model for discovering new uses of marketed drugs. The MBiRW model developed a new similarity calculation method for constructing a three-layer drug-disease heterogeneous network based on historical drug-disease interaction information, ancillary information between drugs and ancillary information between diseases. New drug-disease interaction associations were then inferred by exploring the heterogeneous network with a Bi-Random Walk algorithm. Chen et al. \cite{n4} novelly used transfer learning to combine drug-disease association domains and drug-target association domains to jointly solve for potential relationships between drugs, diseases, and targets. Since there are many shared drugs between the two domains, this allows information migration between the two domains and alleviates the problem of data coefficients between the respective domains. \par

On the other hand, the computational drug repositioning problem can be seen as an application scenario in the field of recommendation systems. Currently, more and more researchers are understanding the drug discovery problem from the perspective of recommendation systems, where strategies based on matrix factorization models become the dominant approach nowadays. The main idea of matrix factorization-based model is to represent the latent factor of drugs and diseases using two low-dimensional dense vectors, and subsequently compute the similarity between drugs and diseases using inner product operations. Luo et al. \cite{m1} proposed the drug repositioning recommendation system (DRRS) model based on the matrix factorization. The model first constructs a heterogeneous adjacency matrix, which includes drug-disease interaction information, auxiliary information between drugs and auxiliary information between diseases. The DRRS model then uses a fast singular value threshold algorithm to complement the unknown positions in the adjacency matrix to infer the potential therapeutic diseases of the drug. To improve the learning ability of the inner product operation and overcome the data sparsity problem, Yang et al. \cite{m2} also borrowed the idea of the matrix factorization model and proposed the Additional Neural Matrix Factorization model (ANMF). The ANMF model extracts the latent factor of drugs and diseases using a variant of autoencoder, which combines the drug-disease interaction information and the respective auxiliary information to overcome the data sparsity. In addition, the ANMF model improves its learning capability by improving the inner product operation using neural networks. To enable the model to take into account the role of partial drug-disease strong associations or local structural information, He et al. \cite{m3} proposed the Hybrid attentional memory network (HAMN) based on the ANMF model, which utilizes memory units and attentional mechanisms to load the information of drug-disease strong associations into the ANMF model. This enables it to consider both global structural information and local structural information of drug-disease associations, thus substantially improving the generalization ability on new data. Yang et al. \cite{m4} proposed multi-similarities bilinear matrix factorization (MSBMF) for drug rediscovery. The novelty of this method is that it can dynamically process the similarity information of multiple drugs and diseases simultaneously, and in addition, it maintains the similarity information and association information to be processed simultaneously during the training process. Finally, MSBMF uses a non-negative function restriction, which makes the variation interval of the final output value conform to common sense. \par

However, two limitations arise from the application of matrix factorization model \cite{mf} in the field of computational drug repositioning. The first limitation is that the matrix factorization model performs the inner product \cite{inner,ncf} on the latent factor vectors of the drug and the disease to derive the predicted values. However, the inner product only considers the angles between the vectors and the respective magnitudes, which is essentially a similarity measure if the magnitudes of the vectors are not considered. The drawback of the similarity measure is that it does not conform to the triangle inequality \cite{angle}, i.e., " the distance between two points cannot be larger than the sum of their distances from a third point", thus limiting the expressiveness of the matrix factorization model to capture the flexible drug-disease associations. A specific example will be presented in Section 2.3.1. To compensate for the deficiency in the expressiveness of the inner product, inspired by the Euclidean distance \cite{cmf} (consistent with the triangular inequality), we consider the latent factor vector of drugs and diseases as a point in the high-dimensional coordinate system and propose a new generalized distance to measure the association between drugs and diseases. However, another limitation of the matrix factorization model emerges in the form of the high-dimensional coordinate system representation \cite{metricf}. The traditional matrix factorization model only considers drug-disease associations in the training parameter phase, which leads to the distance between drugs and diseases with treatment associations being made closer in the high-dimensional coordinate system, while the distance between drugs and diseases without treatment associations being made farther away. However, based on the benchmark principle that similar drugs can have similar diseases, taking drugs as an example, the distance of similar drugs in the high-dimensional coordinate system should be close enough, and the distance of dissimilar drugs should be farther. Obviously, the traditional matrix factorization model cannot achieve the above-mentioned effects. The specific examples will be presented in Section 2.3.2. \par

Therefore, to overcome the aforementioned limitations of existing works, we have proposed a novel computational drug repositioning model, Neural Metric Factorization (NMF), which is used to mine new drug-disease associations. To compensate for the deficiency in the expressiveness of the inner product, the NMF model views the latent factor vector of drugs and diseases as a point, rather than a vector, in the high-dimensional coordinate system, and novelly uses a data-driven generalized Euclidean distance to represent the association between drugs and diseases. That is, the distance between a drug and a disease is closer where a therapeutic relationship exists, and vice versa. This generalized Euclidean distance differs from the Euclidean distance in that its weight values for each dimension of the latent factor of drugs (diseases) are not 1, but are learned from the data. Its connection with the Euclidean distance is that both are equivalent when the weight parameter of all dimensions is 1. Thus Euclidean distance is essentially a special version of generalized Euclidean distance. In addition, to make the distances of the latent factor vectors between similar drugs (diseases) in the high-dimensional coordinate system consistent with the above-mentioned assumptions, the NMF model embeds two metrics information, such as the therapeutic behavior of a drug and the structured similar information between drugs, into the encoding space of its latent factor vectors, so that it can make the distances of the latent factor vectors of similar drugs vectors are close enough in the high-dimensional coordinate system. The operation of extracting the latent factor vector of the disease is similar to that of the drug. \par

The main contributions made by this work can be summarized in the following three points.

\begin{enumerate}
	\item We view the latent factor vector of drugs (diseases) as a point and propose a new generalized Euclidean distance for replacing the inner product in the matrix factorization model to enhance its expressiveness on unknown data.
	\item We novelly embed multiple metric information of drugs (diseases) into the encoding space of latent factor vectors to ensure that similar drugs (diseases) are close enough in the high-dimensional coordinate system.
	\item We run a number of relevant experiments on two real datasets to demonstrate the effectiveness of the above improvement points, as well as the superiority of the NMF model.
\end{enumerate}

The follow-up of this work is organized as follows. Section 2 focuses on the definition of the computational drug repositioning problem, the relevant data sets, the specific examples of two defects of the matrix factorization model and the specific implementation of the neural metric factorization model. Subsequently, Section 3 will focus on the hyperparametric experiments of the NMF model and the validation of the NMF model. Section 4 will summarize the content of this work and the outlook of future work. \par

\section{Materials and Methods}

In this study, we propose a novel neural metric factorization algorithm for discovering new uses of marketed drugs. Firstly, we will briefly introduce the benchmark datasets used in this work. Secondly, we will describe the computational drug repositioning problem from the perspective of the matrix factorization model. Then, we will demonstrate two major limitations of matrix factorization models with concrete examples. Finally, we will introduce the specifics of the NMF model. \par

\subsection{Datasets}

This study uses two benchmark datasets, Gottlieb dataset and Cdataset \cite{m1}. The Gottlieb data set includes 593 marketed drugs, 313 known diseases, and 1933 drug-disease associations that have been verified. Cdataset includes 663 marketed drugs, 409 known diseases, and 2353 drug-disease associations that have been verified. All drug-specific information in the above two datasets is sourced from the DrugBank database \cite{drugbank}, which contains a large amount of experimentally validated drug and target information. Diseases are sourced from the Online Mendelian Inheritance in Man (OMIM) database \cite{omim}, which provides a large amount of publicly available information on human diseases. \par

In addition, the similarity information between drugs \cite{cdk,smile} or diseases for both datasets was also obtained from the existing work \cite{m1}. The similarity values range from 0 to 1, with larger values representing greater similarity and vice versa. The detailed statistics for both the Gottlieb dataset and Cdataset are shown in Table 1. \par

\begin{table}[t]
	\caption{The detailed statistics for two datasets}
	\centering
	\setlength{\tabcolsep}{7.5mm}
	\begin{tabular}{@{}ccccc@{}}
		\toprule
		Datasets & Drugs & Diseases & Validated Associations & Sparsity \\ \midrule
		Gottlieb & 593   & 313      & 1933                   & 0.0104   \\
		Cdataset & 663   & 409      & 2532                   & 0.0093   \\ \bottomrule
	\end{tabular}
\end{table}

\begin{figure}[h]
	\centering
	\includegraphics[scale=0.8]{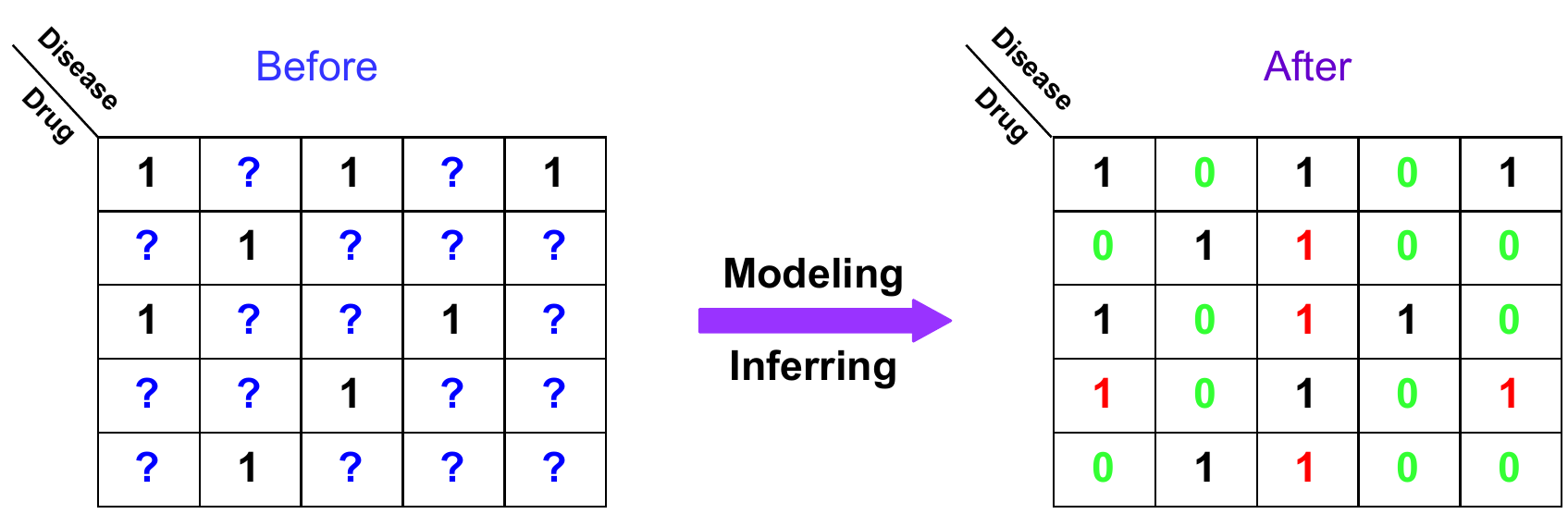}
	\caption{The schematic diagram of computational drug repositioning from the perspective of matrix factorization.}
	\label{}
\end{figure} \par

\subsection{Definition of the computational drug repositioning problem}

Figure 1 shows a schematic diagram of computational drug repositioning from the perspective of matrix factorization model. Under the idea of the matrix factorization model, computational drug repositioning can be equivalent to a matrix completion problem. The left side of Figure 1 shows the original drug-disease association matrix, where the rows represent drugs and the columns represent diseases. When the value of $R_{ij}$ is ‘1’, it means that there is a validated therapeutic association between the $i$th drug and the $j$th disease. Similarly, if the value of $R_{ij}$ is ‘?’ indicates that there is no known therapeutic association between the drug and the disease. The prediction model then uses the known treatment relationship to train the weight parameters it contains so that it can predict the scores of the blank values in the association matrix. The right-hand side of Figure 1 shows the complemented drug-disease association matrix, where '1' in red font and '0' in green font are the probability values predicted by the model, indicating the presence or absence of a treatment relationship between the $i$th drug and the $j$th disease, respectively. \par

\subsection{Two major limitations of the matrix factorization model}

\begin{figure}[h]
	\centering
	\includegraphics[scale=0.97]{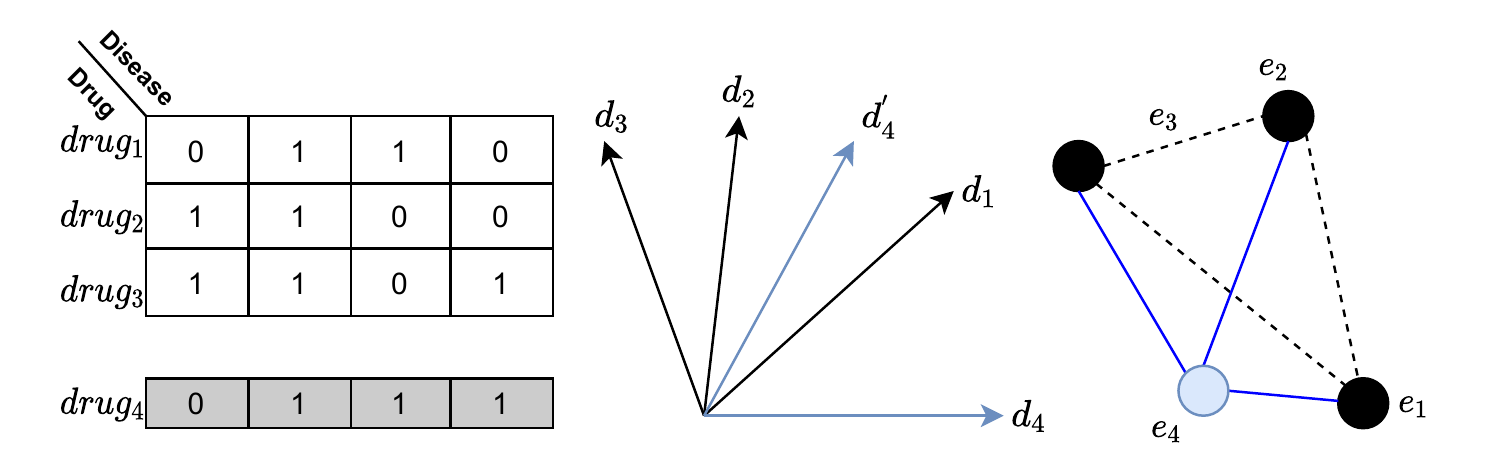}
	\caption{An example explaining the limitations of the inner product in terms of expressiveness; Left: Drug-disease association matrix; Middle: The lantent factor vector of the drugs; Right: Drug representation using points and distances.}
	\label{}
\end{figure} \par

\subsubsection{Inner product}

The mathematical formula of the inner product in matrix factorization is equivalent to the product of cosine and module between two vectors. If the value of module of all vectors in the space is 1 by default, the result of the inner product between two vectors is equal to the cosine of their angle. So the inner product operation is essentially a measure of similarity rather than distance. \par
Figure 2 explains the limitations of the inner product in terms of expressiveness. The therapeutic behavior of the drug for the disease is equivalent to the similarity information between drugs, and we calculate the similarity between drugs from the drug-disease association matrix using the Jaccard similarity measure. First we consider three drugs, $drug_1$, $drug_2$ and $drug_3$. Their similarity priorities are $s_{23}$>$s_{12}$>$s_{13}$. Therefore, their expressions in the latent factor space can be expressed as $d_1$,$d_2$ and $d_3$. The angle between these three vectors corresponds to the similarity information of the drugs on the interaction matrix, i.e., the more similar the drugs are the smaller the angle between them. However, if a fourth drug $drug_4$ is now added, its similarity information on the interaction matrix is, $s_{41}$>$s_{43}$>$_{42}$. But we could not find a suitable $d_4$ in the latent factor space to satisfy the above similarity restriction. If we consider the latent factor of a drug as a point in a high-dimensional coordinate system and express the similarity between drugs in terms of distance, it is easy to find a suitable representation as shown in $e$, where the more similar the drug is the closer it is and vice versa.

\begin{figure}[h]
	\centering
	\includegraphics[scale=0.91]{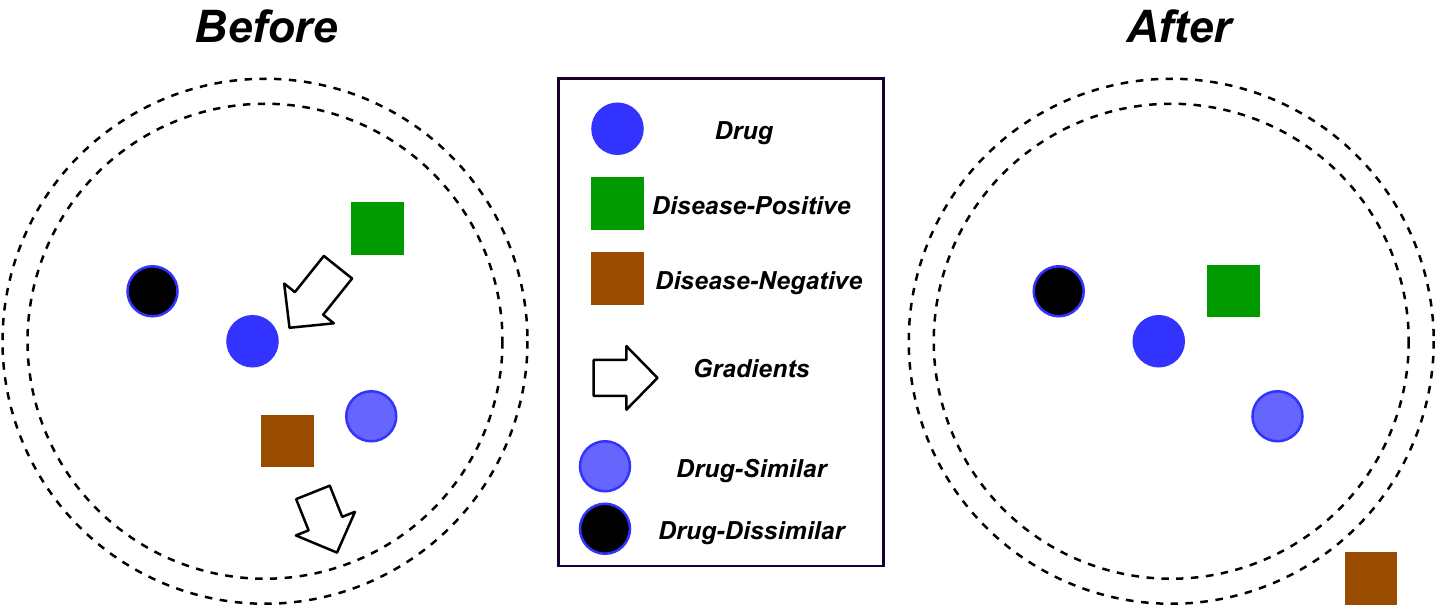}
	\caption{An example explaining the limitations of the encoding space of latent factor vectors}
	\label{}
\end{figure} \par

\subsubsection{Encoding space of Latent factor vectors}

However, another limitation of the matrix factorization model emerges in this form of expression of points and distances. Figure 3 shows the change of latent factor positions of drugs and diseases in the traditional matrix factorization model before and after training. The circles in the figure represent drugs and the boxes represent diseases. Because the matrix factorization model only considers the validated drug-disease treatment associations during training process, this results in a gradient between drugs and diseases with associations, so that the distance between the target drug and the disease will be closer as the parameters change, and the distance between the target drug and the disease without validated associations will be farther. \par

However, the position of the latent factor of the remaining drugs in the left side of Figure 3 did not change. This is because there is no other metric information available for the latent factor between drugs, resulting in no gradient in the parameter update. One of the major assumptions in the computational drug repositioning domain is that "similar drugs usually treat similar diseases", therefore, the position of the latent factor of each drug should show the similarity information between them, i.e., the distance of the latent factor of similar drugs should be close enough, and the distance of the latent factor of dissimilar drugs should be distant.  \par

\begin{figure}[h]
	\centering
	\includegraphics[scale=0.75]{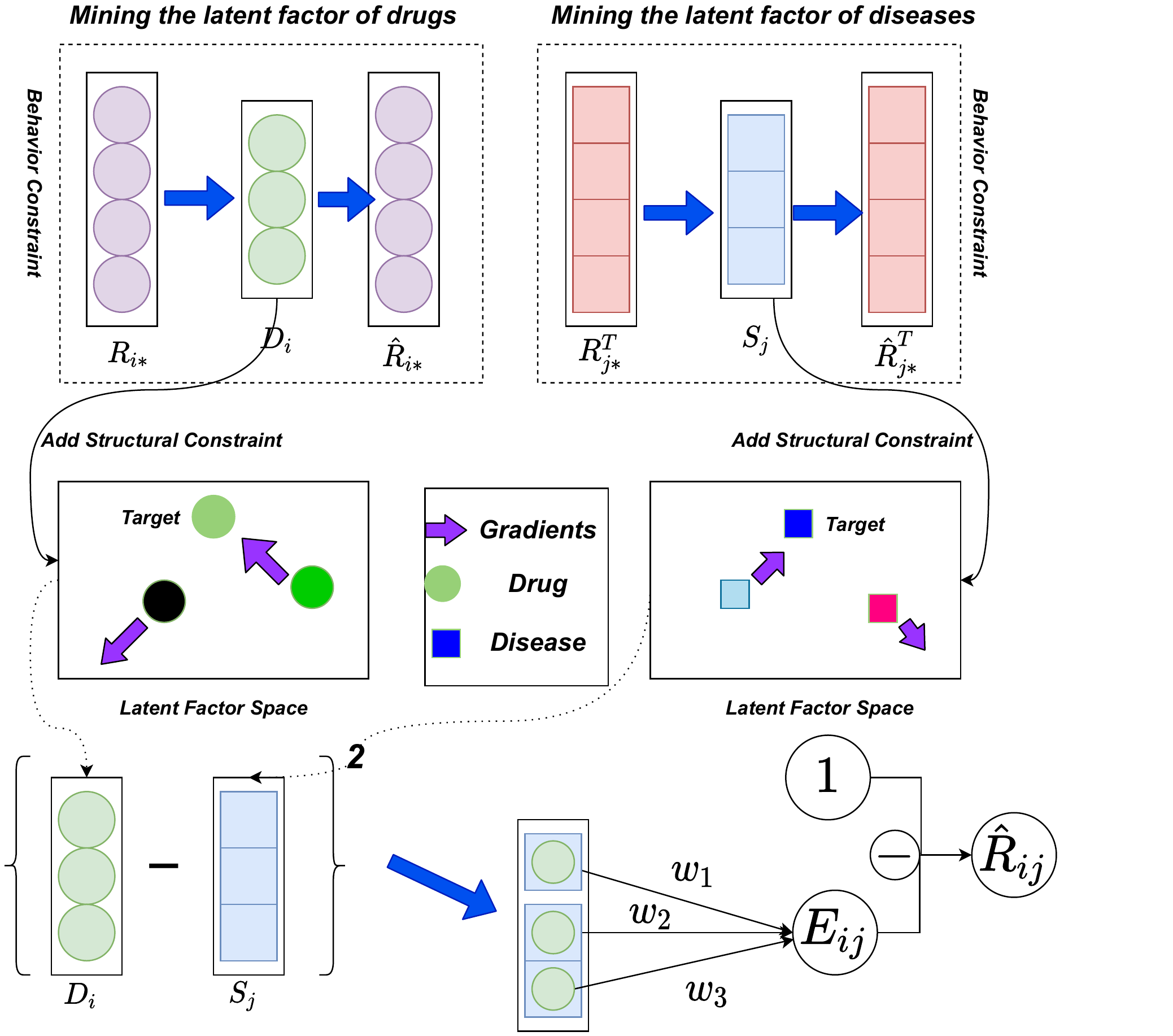}
	\caption{The framework of the NMF model.}
	\label{}
\end{figure} \par

\subsection{Neural Metric Factorization}

To address the above two shortcomings of the matrix factorization model, we novelly propose the neural metric factorization (NMF) algorithm in this work. The NMF model has unique improvements for each limitation, including the use of distance metric instead of inner product in section 2.4.1, and the use of metric information to optimize the position of the latent factor of drugs and diseases in the high-dimensional coordinate system in section 2.4.2. Figure 4 shows the framework of the NMF model.\par

\subsubsection{Inferring drug-disease associations using point and generalized Euclidean distance}

In order to compensate for the deficiency in the expressiveness of the inner product, unlike the matrix factorization model that uses latent factor vectors to represent drugs and diseases, the NMF model treats latent factor vectors as points in the high-dimensional coordinate system to represent drugs and diseases. And we propose a new generalized Euclidean distance for measuring the association between drugs and diseases. The NMF model uses two latent factor matrixes $D$ and $S$ to represent drugs and diseases, to determine their respective point positions in the high-dimensional coordinate system. Where $d_i$ is the latent factor vector of the $i$th drug and $s_j$ is the latent factor vector of the $j$th disease. \par

Then, to better represent the distance relationship between drugs and diseases in the high-dimensional coordinate system, we introduced a generalized Euclidean distance formula as shown in Equation (1) based on Euclidean distance and neural network. The generalized Euclidean distance formula differs from the traditional Euclidean distance in that it is a data-driven distance metric rather than a heuristic. It is equipped with a learnable weight parameter in each dimension of the latent factor vector to adjust the percentage of each dimension in the final output value. \par

\begin{equation}
E_{ij} = \sum_{t=0}^k {w_t(d_{it}-s_{jt})^2}
\end{equation}

Where $ E_{ij} $ denotes the distance between drug $i$ and disease $j$ in the high-dimensional coordinate system. The closer the distance is, the higher the probability of a therapeutic relationship between them and vice versa. However, considering that the label used in the training process of the NMF model is the drug-disease association value $R_{ij}$, which is essentially the value under the similarity measure between the drug and the disease. Distance and similarity are two opposite concepts, where a large value of similarity implies a smaller value of the distance. Therefore, in order to unify the concepts, we use the conversion formula shown in Equation (2) to convert the distance metric to the similarity metric. \par

\begin{equation}
\hat{R}_{ij} = 1- \frac{1}{1+e^{E_{ij}}}
\end{equation}

By the above conversion, the distance between the drug and the disease can be smoothly converted into the treatment probability. The larger the value of $ E_{ij} $, the smaller the value of $\hat{R}_{ij}$, and vice versa. $e$ is the Euler's number. $\hat{R}_{ij}$ is the probability of a therapeutic relationship between drug $i$ and disease $j$ as inferred by the NMF model. \par

\subsubsection{Extracting latent factor using metric information}

However, following the traditional idea of matrix factorization model, only the validated drug-disease associations are considered during the parameter training process. This results in the final values of the two latent factor matrices $D$ and $S$ not satisfying the restriction mentioned in Section 2.3.2 that similar drugs should be located as close as possible to each other in the high-dimensional coordinate system, and the same goes for diseases. \par

Therefore, in order to make the values of the latent factor matrix $D$ and $S$ meet the above conditions,  taking drugs as an example (diseases are similar to them), we embed the metric information such as therapeutic behavior records and structural similarity information of drugs to the encoding space of the latent factor so that it can restrict the variation interval of the latent factor of drugs. The calculation procedure of mining the latent factor of the drug using metric information is shown in Equations (3)-(5). \par

\begin{align}
d_{i} &= f(W_1 R_{i*} + b_1) \\ 
\hat{R}_{i*} &= g(V_1 d_i + b_d) \\ 
Loss_{d} &= \left \|   \hat{R}_{i*} - R_{i*}  \right \|^2 + \sum_{k} DrugSim_{ik} \left \|   d_{i} - d_{k}  \right \|^2 
\end{align}

The intuition behind the above equations is as follows. Firstly, the input $R_{i*}$ in equation (3) is the therapeutic behavior of drug $i$ for all diseases. Similar drugs will usually have similar therapeutic behavior. Letting this information be the input will cause drugs with similar therapeutic behavior to obtain a similar latent factor in the subsequent computation. Equation (4) is the operation of reconstructing the input using the latent factor since the effectiveness of the latent factor depends on its ability to completely restore the input $R_{i*}$. In addition, the loss function shown in Equation (5), not only the error between the input and reconstruction input is included, but we also add the error between the latent factor of the neighboring drug $k$ and the target drug $i$, and the weight parameter $DrugSim_{ik}$ before this error is the structural similarity between the neighboring drug $k$ and the target drug $i$. If a large value of $DrugSim_{ik}$ indicates that the similarity between drugs $i$ and $k$ is large, then in order to minimize the loss function, the model makes the error values of the latent factor of drugs $i$ and $k$ as small as possible, i.e., the points in the high-dimensional coordinate system are as close as possible. \par

\begin{figure}[h]
	\centering
	\includegraphics[scale=0.91]{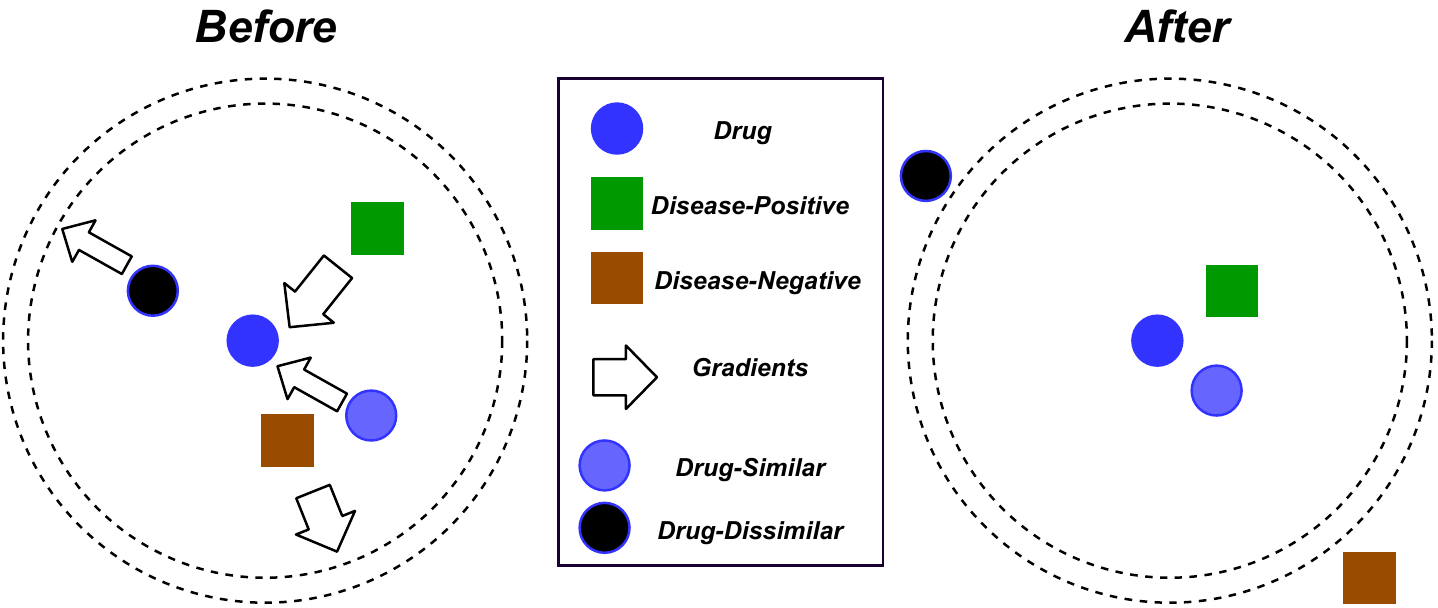}
	\caption{The process of change in the position of the latent factor of drugs and diseases.}
	\label{}
\end{figure} \par

Figure 5 shows the changing process of the position of the latent factor vector of drugs and diseases in the high-dimensional coordinate system after adding the restriction of metric information. We can observe that the distance between similar drugs (diseases) becomes closer and vice versa. \par

The principle of extracting the latent factor of the disease is similar to that of the drug, and the Equations are shown in (6)-(8). \par

\begin{align}
s_{j} &= f(W_2 R^T_{j*} + b_2) \\ 
\hat{R}^T_{j*} &= g(V_2 s_j + b_s) \\ 
Loss_{s} &= \left \|   \hat{R}^T_{j*} - R^T_{j*}  \right \|^2 + \sum_{k} DisSim_{jk} \left \|   s_{j} - s_{k}  \right \|^2 
\end{align}

Where $s_j$ is the latent factor of disease $j$, $ R_{*j}$ is the treated record of the disease for all drugs,$ DisSim$ is the similarity matrix between diseases. It can be found that similar diseases are closer to each other, while dissimilar diseases become farther apart.\par

\subsubsection{Parameter Learning}

The loss function of the NMF model contains three components, the most significant one is the error between the predicted value $\hat{R}_{ij}$ and the label value $R_{ij}$, and the rest is the error arising when extracting the latent factor of the drug and the error arising when extracting the latent factor of the disease, respectively. Therefore, the loss function of the NMF model is shown in Equations (9)-(10). \par

\begin{align}
Loss &= Loss_p + \alpha Loss_{d} + \beta Loss_{s} \\ 
Loss_p &=  R_{ij} log \hat{R}_{ij} + (1-R_{ij}) log(1-\hat{R}_{ij}) 
\end{align}

The $\alpha$ and $\beta$ are weight parameters used to adjust the proportion of the corresponding loss function in Equation (9). To be able to learn the parameters in the NMF model, we use the Adam optimizer in PyTorch to train the NNNF model by minimizing the loss function $Loss$ in a loop iteration. Algorithm 1 shows the pseudo-code of the NMF model. \par

\begin{algorithm}
	\caption{Neural Metric Factorization.} 
	\textbf{Input-1:} Drug-Disease association matrix, $R$\\
	\textbf{Input-2:} Drug-Drug similarity matrix, $DrugSim$\\
	\textbf{Input-2:} Disease-Disease similarity matrix, $DisSim$\\
	\textbf{Output:} Probability of treatment of disease $j$ by drug $i$, $ \hat{R}_{ij}$
	
	\begin{algorithmic}[1]
		\For {$(i,j)\in$ Unknown drug-disease associations}

		\State Extracting latent factor using metric information.
		\State Drug 
		\State \quad $d_{i} \leftarrow  f(W_1 R_{i*} + b_1)$
		\State \quad $\hat{R}_{i*} \leftarrow g(V_1 d_i + b_d)$
		\State \quad $Loss = \left \|   \hat{R}_{i*} - R_{i*}  \right \|^2 + \sum_{k} DrugSim_{ik} \left \|   d_{i} - d_{k}  \right \|^2  $ 
		
		\State Disease
		\State \quad $s_j \leftarrow f(W_2 R^T_{j*} + b_2)$
		\State \quad $\hat{R}^T_{j*} \leftarrow g(V_2 s_j + b_s)$
		\State \quad $Loss \leftarrow  \left \|   \hat{R}^T_{j*} - R^T_{j*}  \right \|^2 + \sum_{k} DisSim_{jk} \left \|   s_{j} - s_{k}  \right \|^2  $ 
		
		\State  Inferring drug-disease associations using point and generalized Euclidean distances.
		\State \quad $ {E}_{ij} \leftarrow \sum_{t=0}^k {w_t(d_{it}-s_{jt})^2}$
		\State \quad $\hat{R}_{ij} \leftarrow 1- \frac{1}{1+e^{E_{ij}}}$
		\State \textbf{Return $\hat{R}_{ij}$}
		\EndFor
	\end{algorithmic} 
\end{algorithm}

\section{Experiments and Discussion}

To verify the excellence of the neural metric factorization proposed in this paper, we run a wide range of relevant experiments to answer the following three research questions. \par

\begin{itemize}
	\item[\textbf{RQ1}] How does the dimension of the latent factor vector, the key hyperparameter in the neural metric factorization model, affect its performance?
	\item[\textbf{RQ2}] Are the proposed generalized Euclidean distance and the two metric information enhance the prediction ability of the NMF model?
	\item[\textbf{RQ3}] Can the performance of our proposed NMF model outperform the state-of-the-art computational drug repositioning algorithms?
\end{itemize}

\subsection{Evaluation Metrics}

For each computational drug repositioning dataset, we slice all known drug-disease associations in a 7:3 ratio, where 70\% of the known associations are used as the training set to train the weight parameters in the neural metric factorization model as well as other benchmark models; the remaining 30\% of the known associations are used as the test set to evaluate the experimental results of each model for numerical comparison. \par

In order to fairly compare the differences among models, we adopt Area Under Curve (AUC), a mainstream evaluation metric for binary classification problem, as one of the metrics. In addition, the two data sets for drug repositioning are unbalanced data, and the performance of the model cannot be fully reflected if only the AUC is used, so we also add another evaluation metric for unbalanced data, Area Under Precision-Recall (AUPR). Combining these two metrics can reflect the prediction performance of the model more comprehensively. \par


\subsection{Hyperparameter Study (RQ1)}

The dimension of the latent factor vector is an important hyperparameter of the NMF model, and the expressiveness of the NMF model is proportional to its size. Figure 6 shows the experimental results of the NMF model with different latent factor vector dimensions, and the variation intervals of latent factor vector dimensions are $[8,16,32,64,128]$. \par

\begin{figure}[htbp]
	\centering
	\subfigure[Comparing Results (Gottlieb)]{
		\includegraphics[width=6cm,height=4cm]{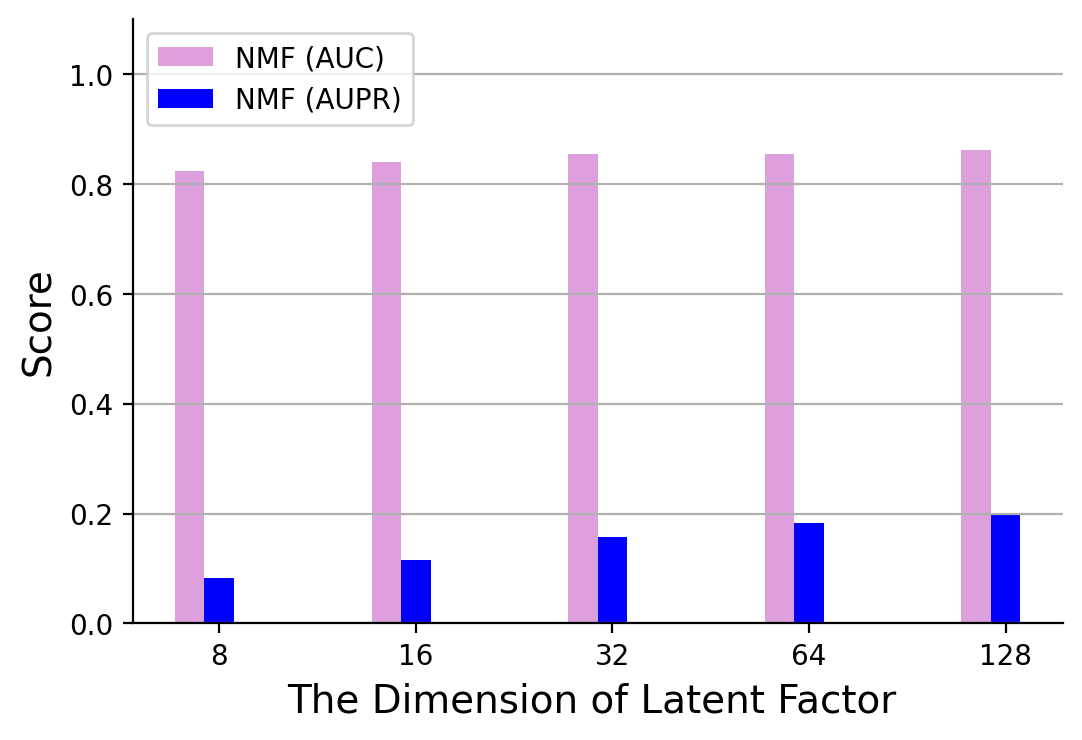}
	}
	\quad
	\subfigure[Comparing Results (Cdataset)]{
		\includegraphics[width=6cm,height=4cm]{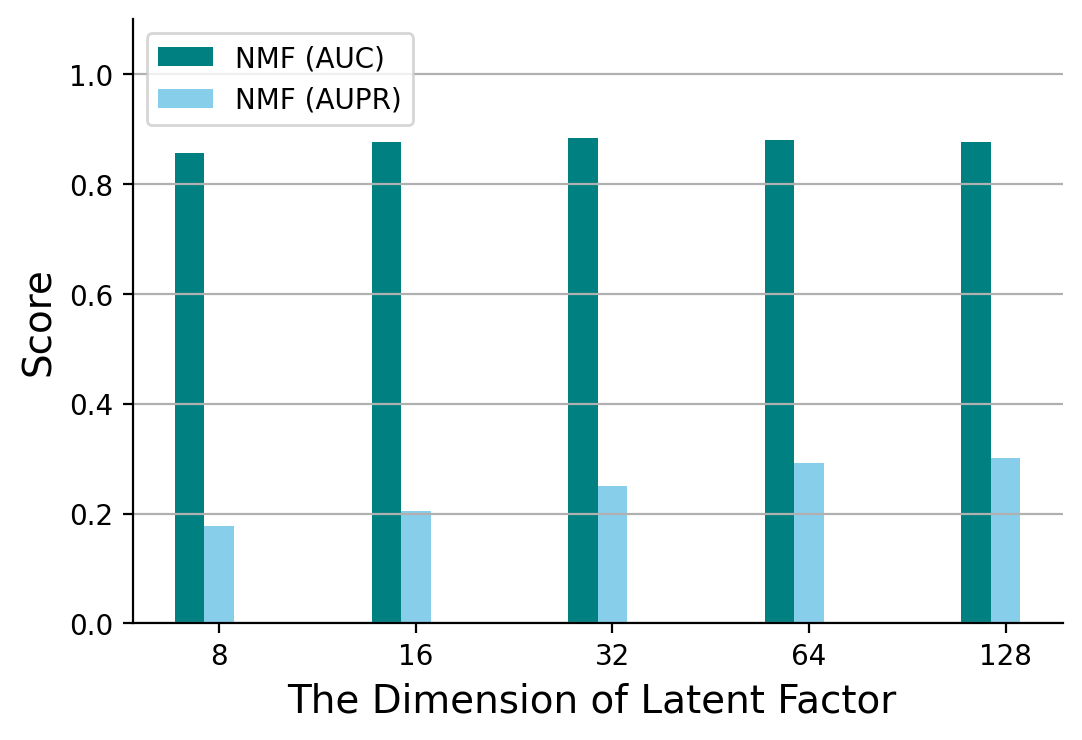}
	}

	\subfigure[ROC Curve (Gottlieb)]{
	\includegraphics[width=6cm,height=4cm]{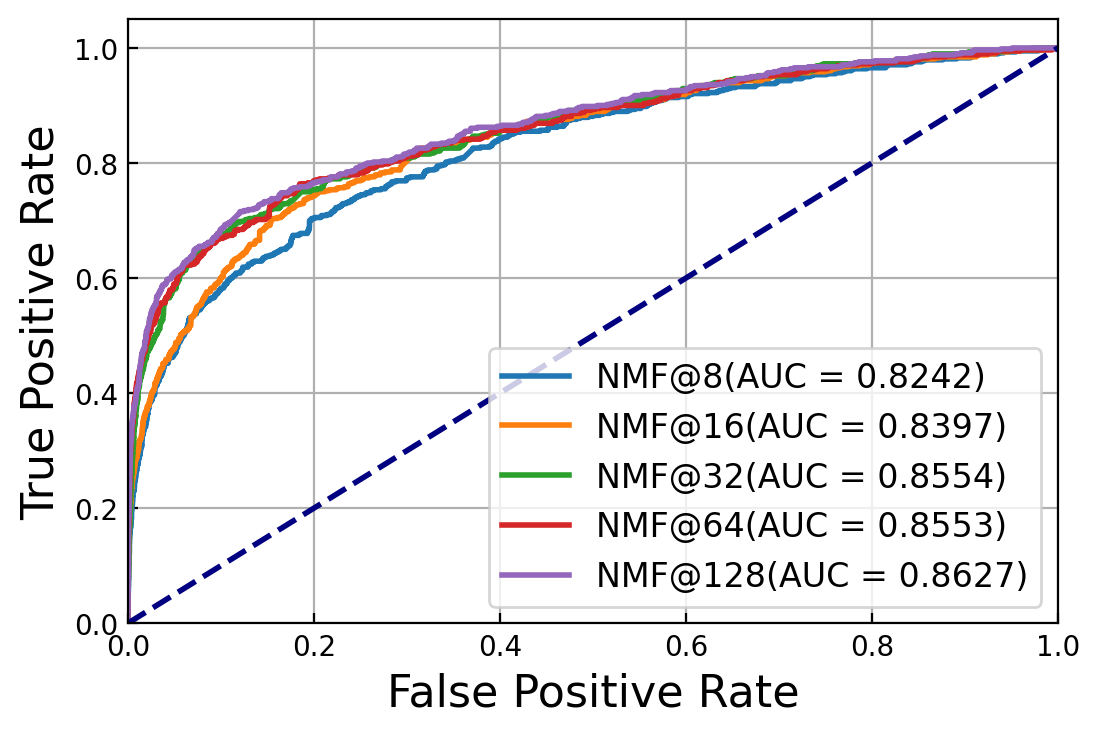}
}
\quad
\subfigure[ROC Curve (Cdataset)]{
	\includegraphics[width=6cm,height=4cm]{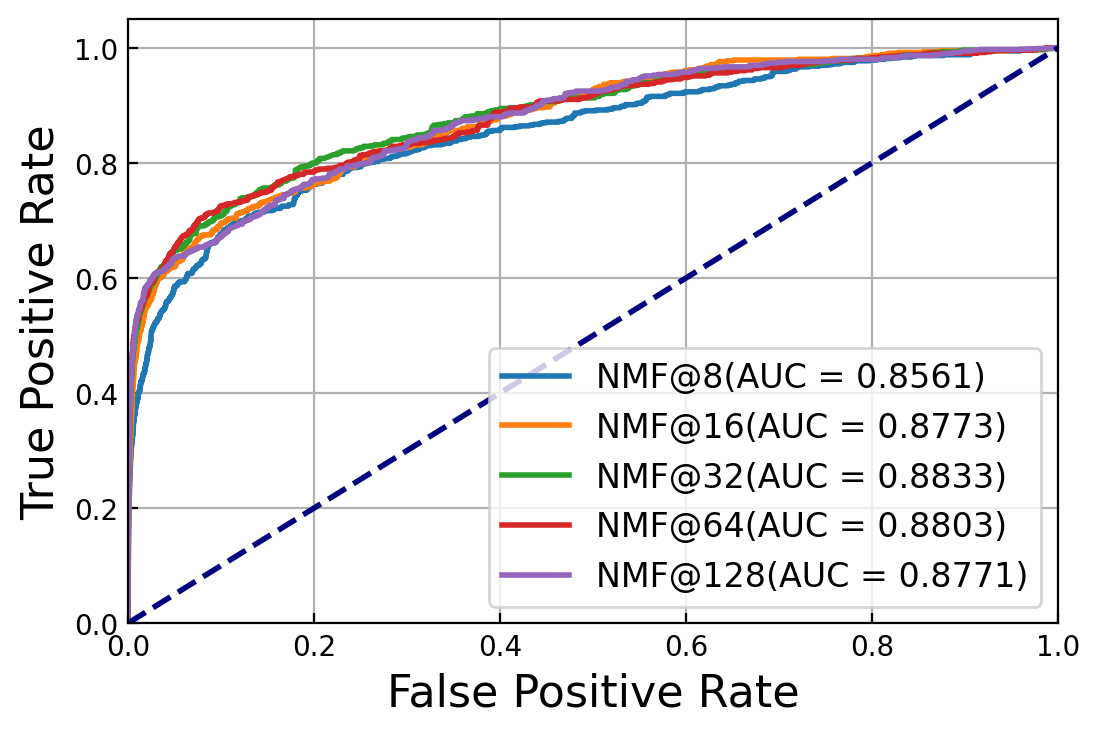}
}

	\subfigure[Precision-Recall Curve (Gottlieb)]{
	\includegraphics[width=6cm,height=4cm]{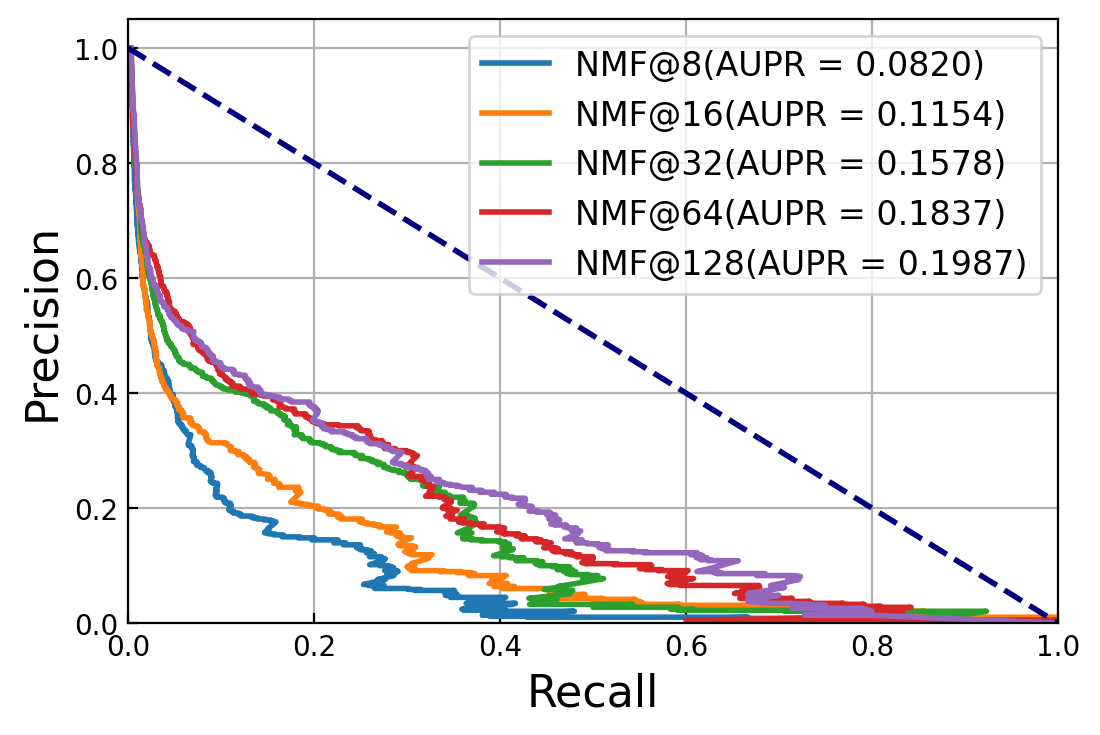}
}
\quad
\subfigure[Precision-Recall Curve (Cdataset)]{
	\includegraphics[width=6cm,height=4cm]{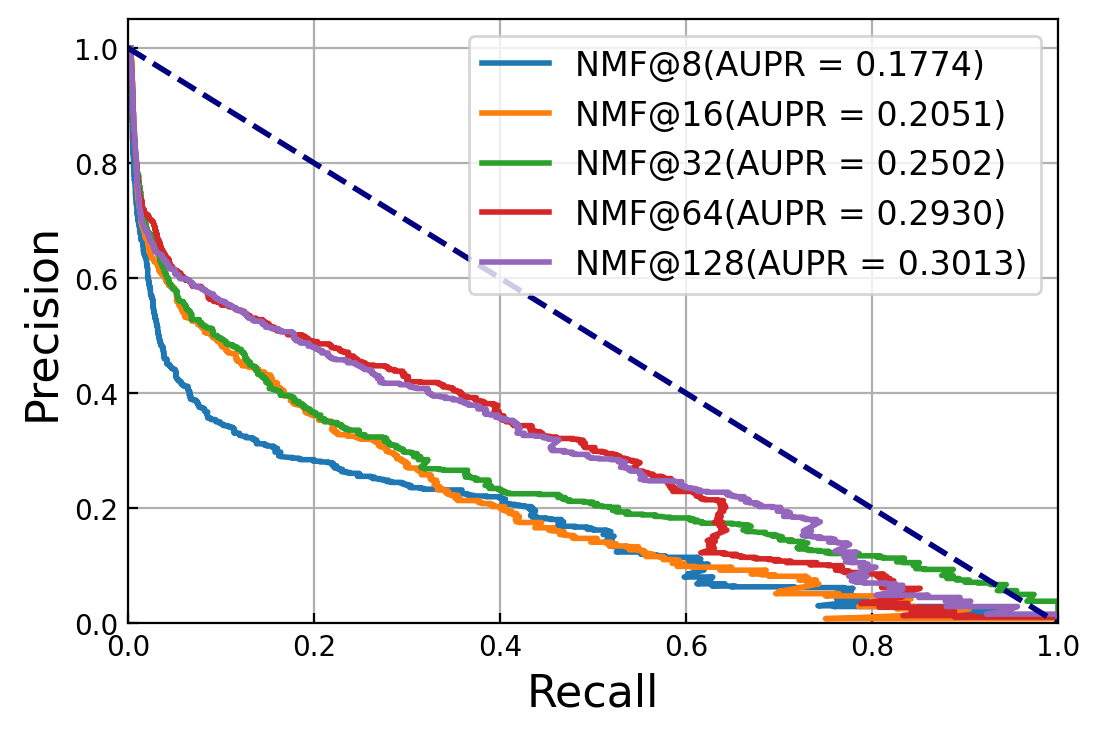}
}
	\centering
	\caption{ The experimental results of NMF model with different latent factor vector dimensions}
\end{figure}\par

We can find that on the Gottlieb dataset, both the AUC and AUPR values of the NMF model become larger with the increase of the vector dimension. However, on the Cdataset, the AUC values reach the maximum value when the dimension of latent factor is 32, and then the effect of the model gradually decreases with the increase of dimensionality. The above phenomenon illustrates that the generalization performance of the NMF model is not proportional to the its complexity, and an overly complex model will misfit the noise in the computational drug relocation dataset, which leads to the overfitting phenomenon. While an overly simple model cannot effectively fit the data distribution in the computational drug repositioning dataset, resulting in an underfitting phenomenon. Therefore, for different datasets, the appropriate dimensionality of the latent factor vector should be trained using the validation set so as to effectively improve the generalization ability of the NMF model. These above experimental analyses above are the answer to \textbf{RQ1}.\par

\subsection{Ablation Study (RQ2)}

\begin{figure}[htbp]
	\centering
	\subfigure[Comparing Results (Gottlieb)]{
		\includegraphics[width=6cm,height=4cm]{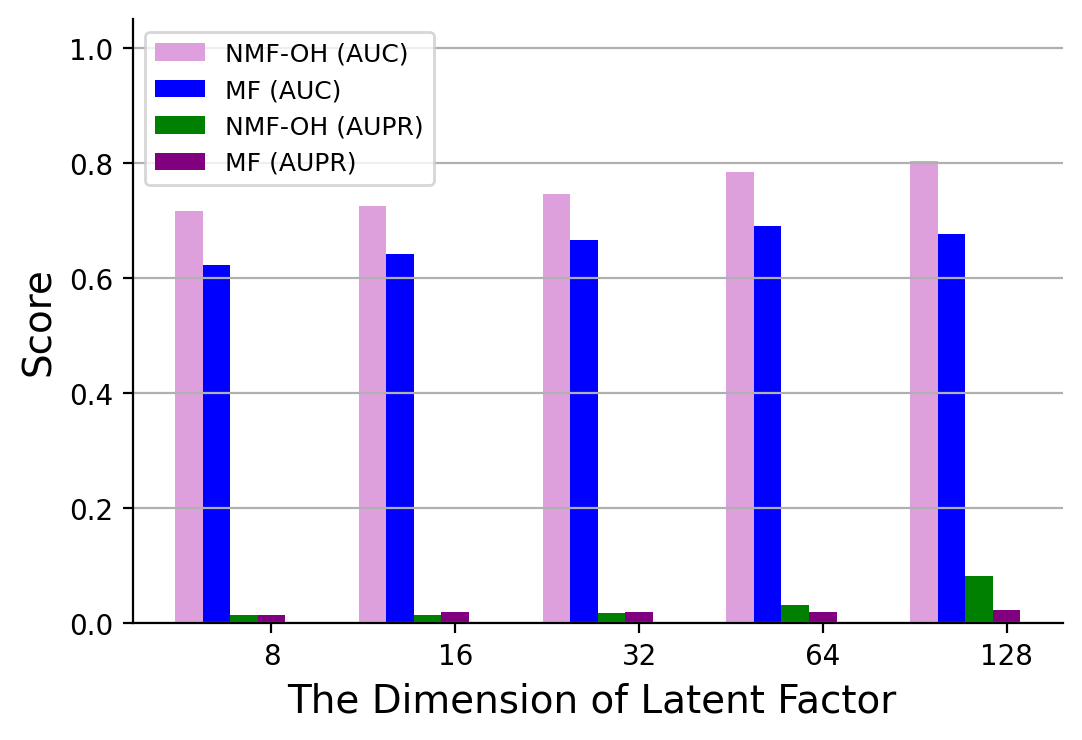}
	}
	\quad
	\subfigure[Comparing Results (Cdataset)]{
		\includegraphics[width=6cm,height=4cm]{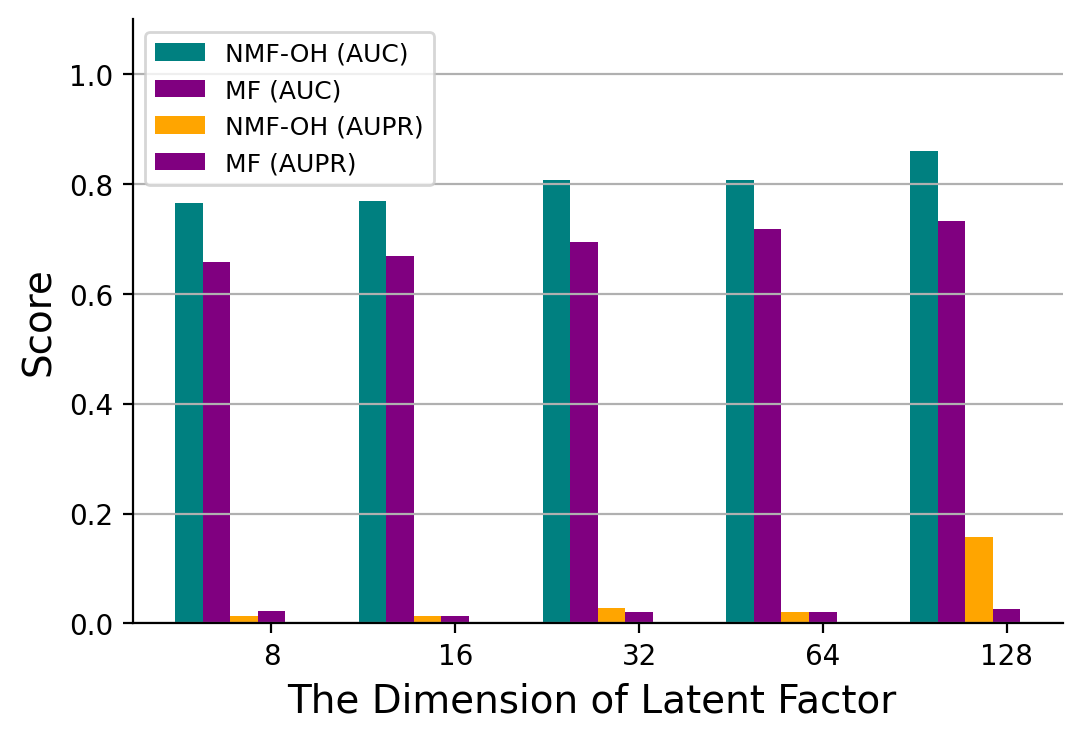}
	}

	\subfigure[ROC Curve (Gottlieb)]{
	\includegraphics[width=6cm,height=4cm]{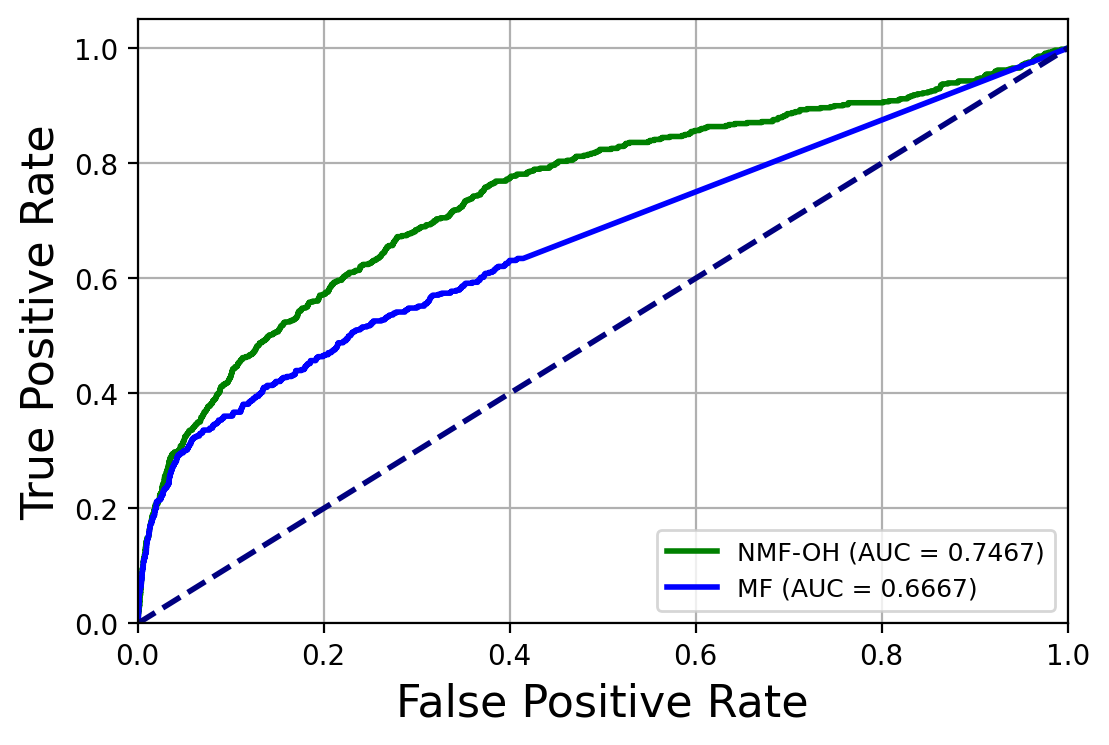}
	}
	\quad
	\subfigure[ROC Curve (Cdataset)]{
	\includegraphics[width=6cm,height=4cm]{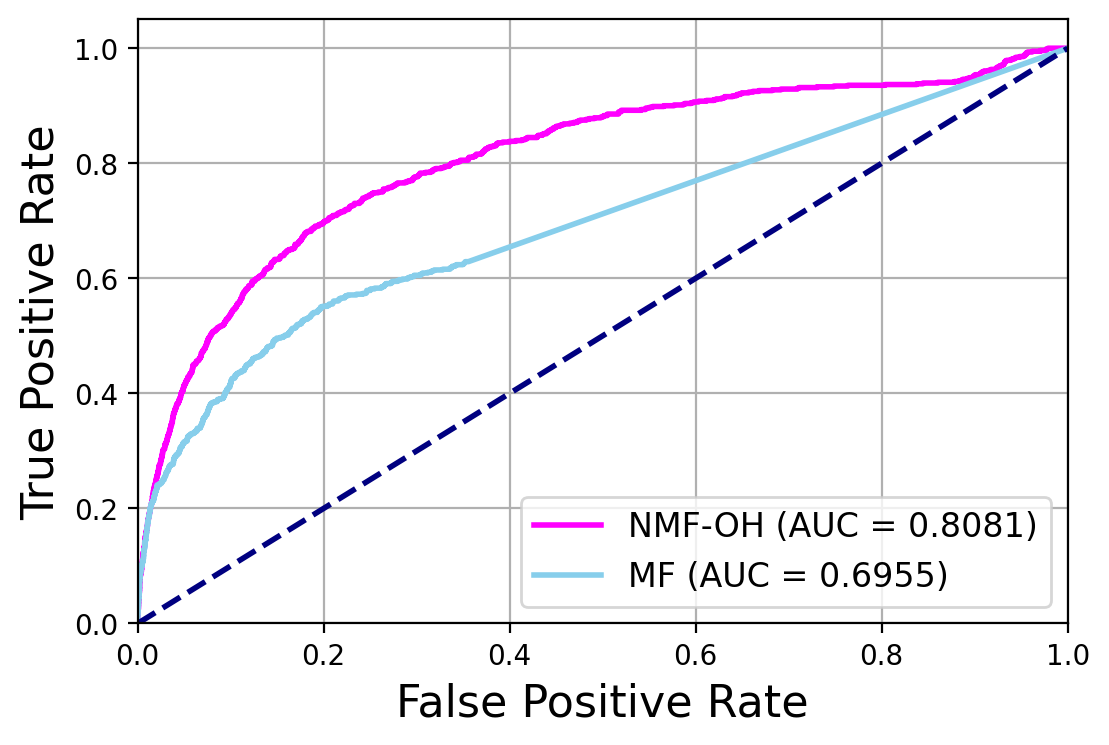}
	}

	\subfigure[Precision-Recall Curve (Gottlieb)]{
	\includegraphics[width=6cm,height=4cm]{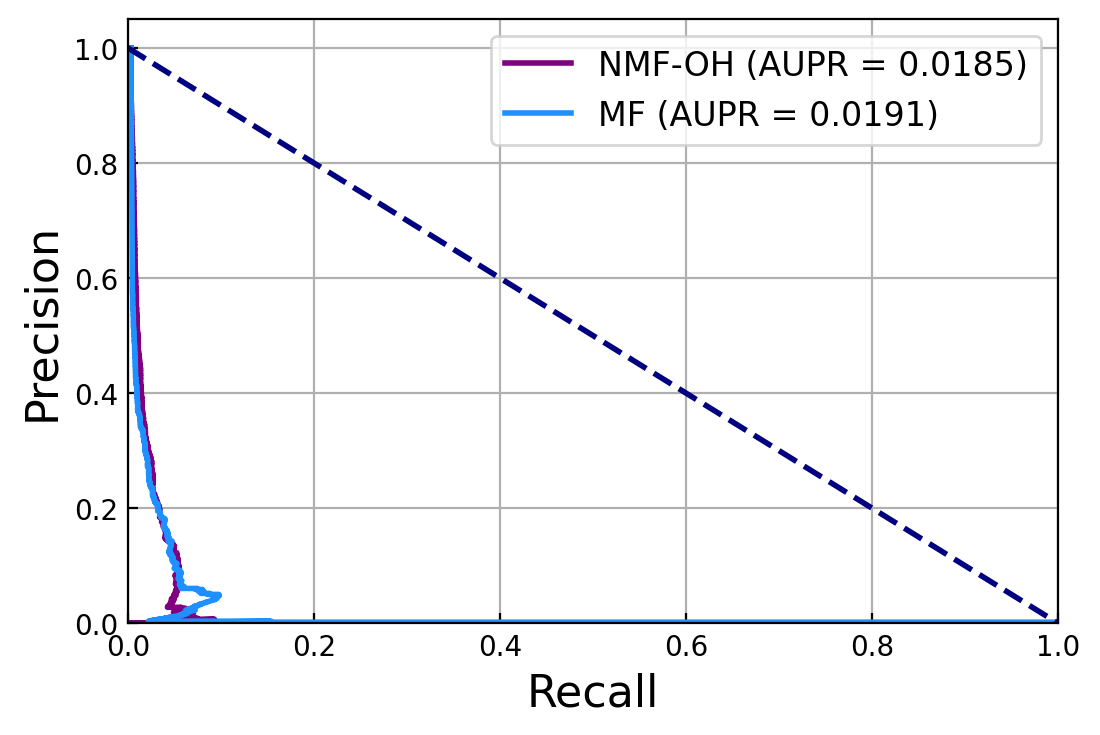}
	}
	\quad
	\subfigure[Precision-Recall Curve (Cdataset)]{
		\includegraphics[width=6cm,height=4cm]{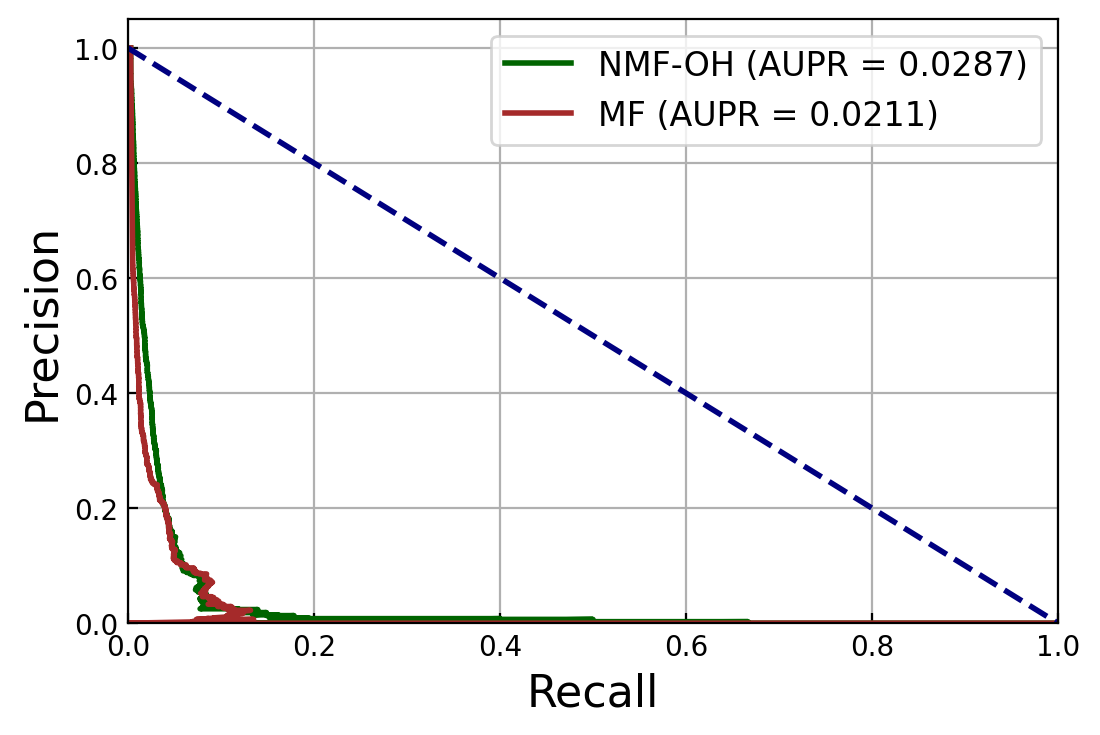}
	}
	\centering
	\caption{ The experimental comparison of neural metric factorization model and matrix factorization model under one-hot coding}
\end{figure}\par

\subsubsection{Validity of Generalized Euclidean Distance}

To be able to show the effectiveness of representing drug-disease associations by points and generalized Euclidean distances, we compare the neural metric factorization model with the traditional matrix factorization model which represents drug-disease associations by vectors and inner product operations. The NMF model does not use the metric information and adopts the same one-hot encoding as the matrix factorization model. In this way, the performance of the generalized Euclidean distance and the inner product operation can be fairly compared. The data set and the remaining hyperparameters used in the comparison model are consistent, and the experimental results of the comparison are shown in Figure 7. \par

In term of AUC value, we can find that the NMF model outperforms the matrix factorization model by a large margin in both datasets, and the effect of NMF-OH was comparable to MF when the dimension of latenr factor was 32 under the AUPR metric. It is noteworthy that NMF model has a greater effect on the AUC metric and is effective in all dimensions of the latent factor vector. In contrast, its improvement effect on the AUPR metric is larger only when the dimensionality of the latent factor vector is larger, which may be due to the large sparsity of the computational drug repositioning dataset. \par

In the case of NMF-OH model with large latent facotr dimension, confirming the positive effect of generalized Euclidean distance on the model performance improvement. \par

\subsubsection{Validity of Metric Information}

\begin{figure}[htbp]

	\subfigure[Comparing Results (Gottlieb)]{
		\includegraphics[width=6cm,height=4cm]{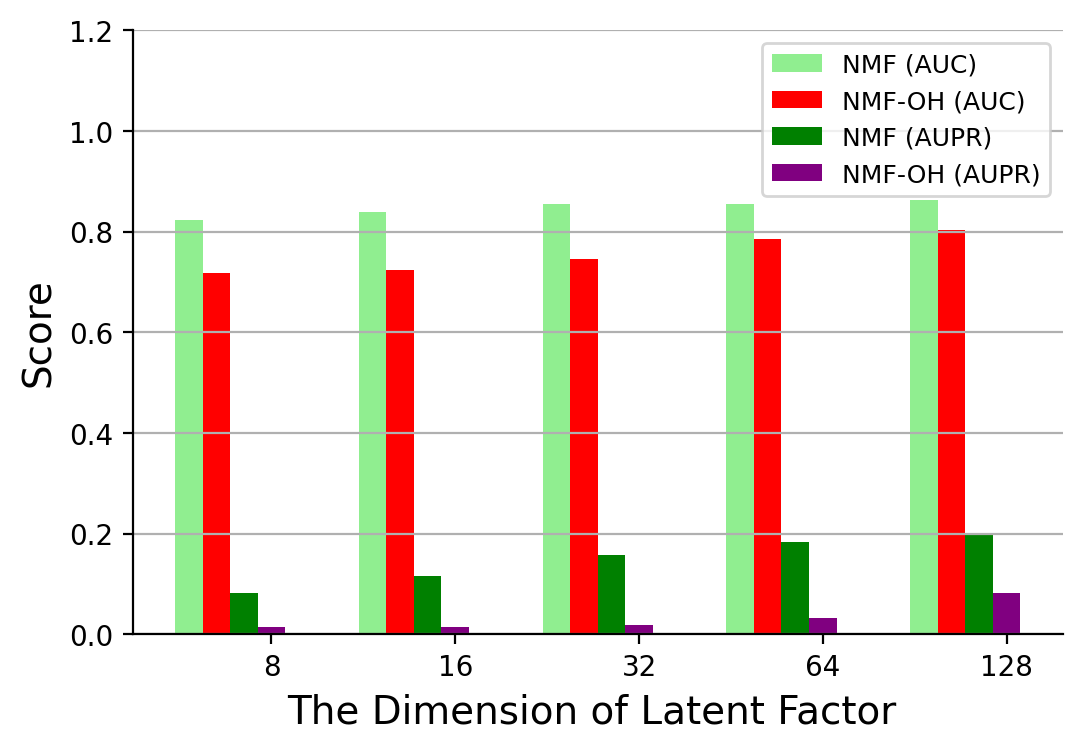}
	}
	\quad
	\subfigure[Comparing Results (Cdataset)]{
		\includegraphics[width=6cm,height=4cm]{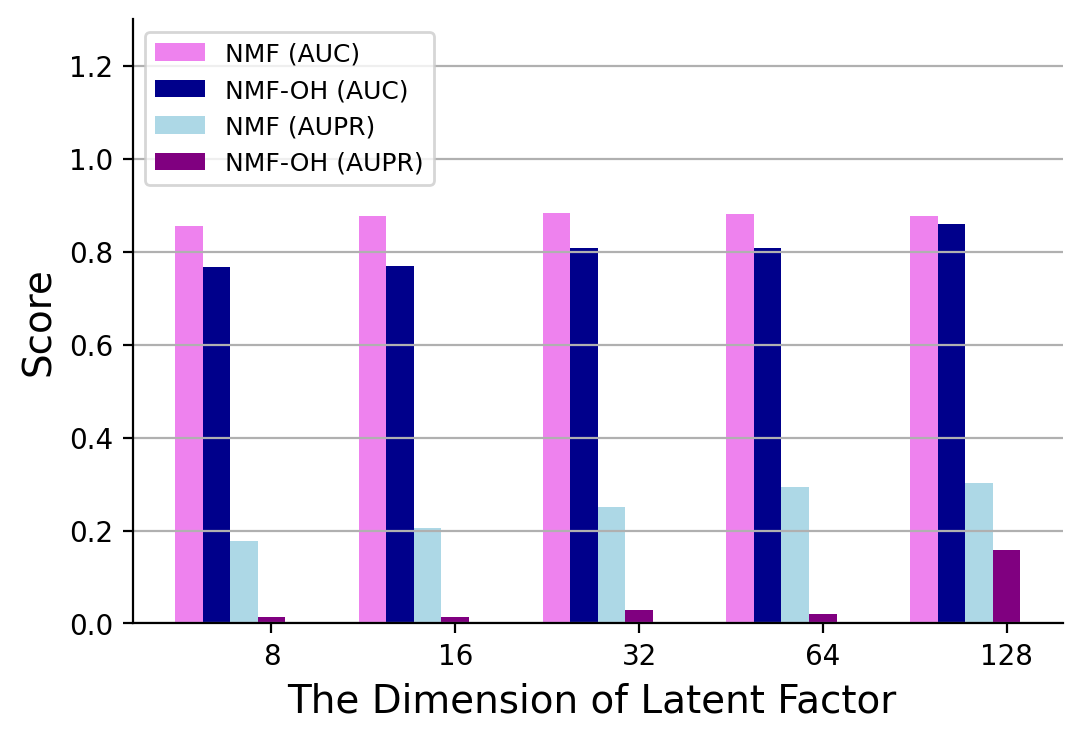}
	}

	\subfigure[ROC Curve (Gottlieb)]{
	\includegraphics[width=6cm,height=4cm]{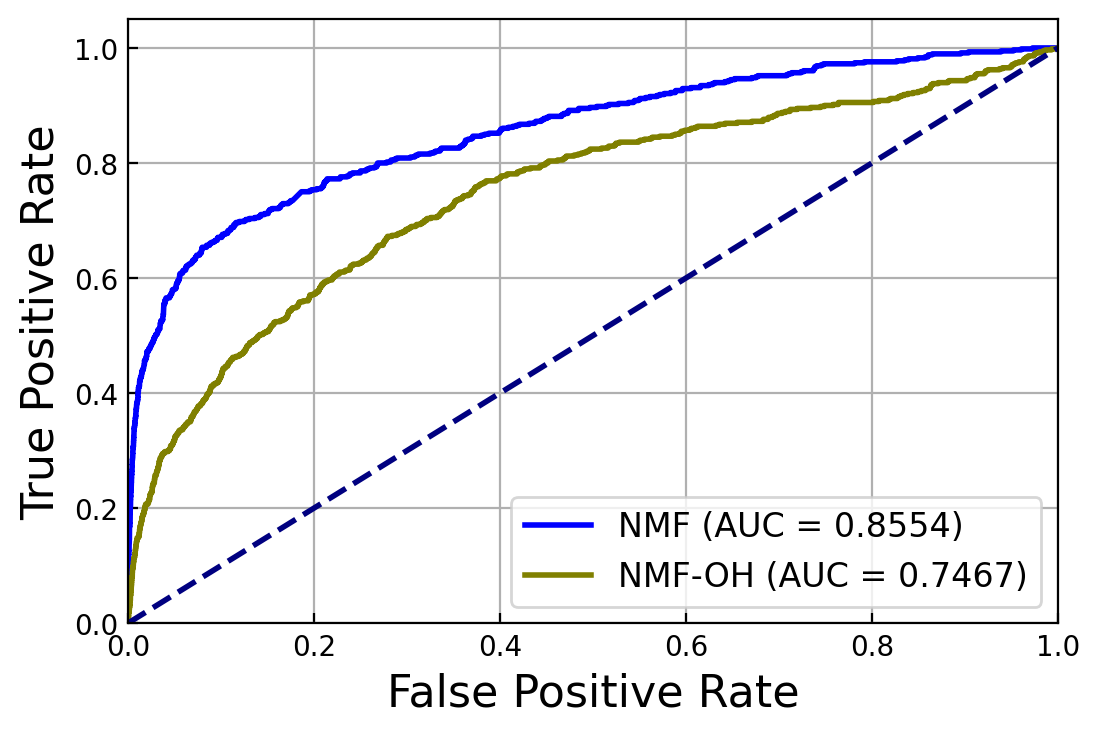}
	}
	\quad
	\subfigure[ROC Curve (Cdataset)]{
		\includegraphics[width=6cm,height=4cm]{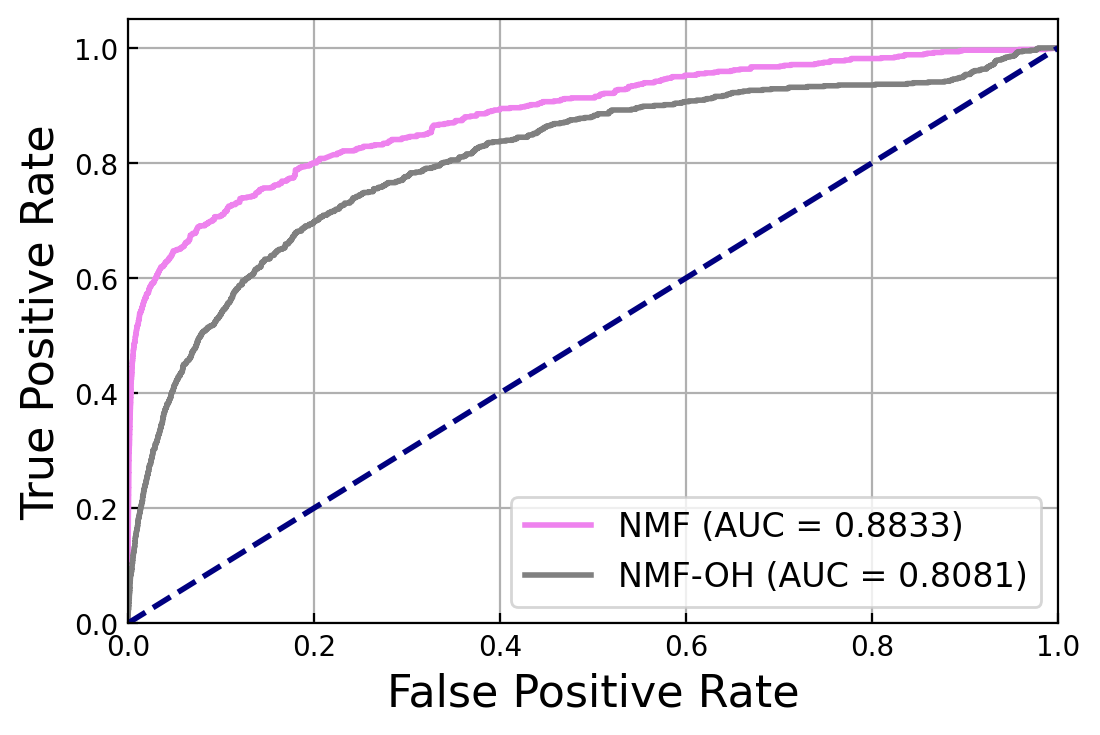}
	}
	\quad	
	\subfigure[Precision-Recall Curve (Gottlieb)]{
		\includegraphics[width=6cm,height=4cm]{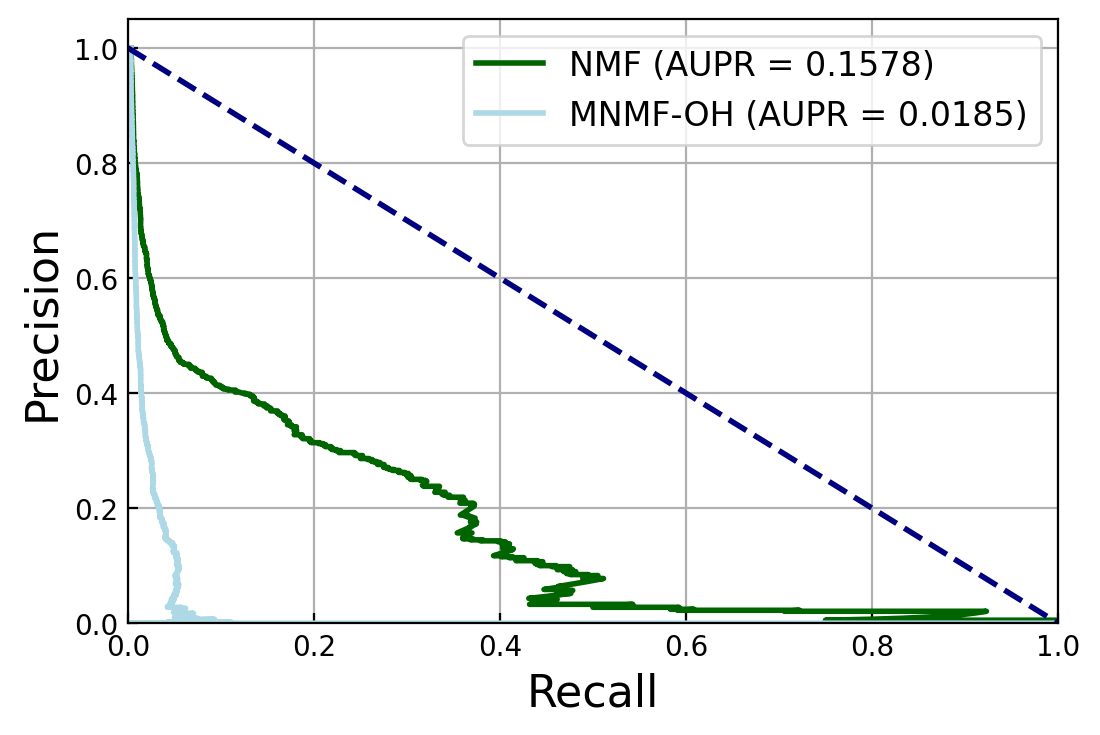}
	}
	\quad
	\subfigure[Precision-Recall Curve (Cdataset)]{
		\includegraphics[width=6cm,height=4cm]{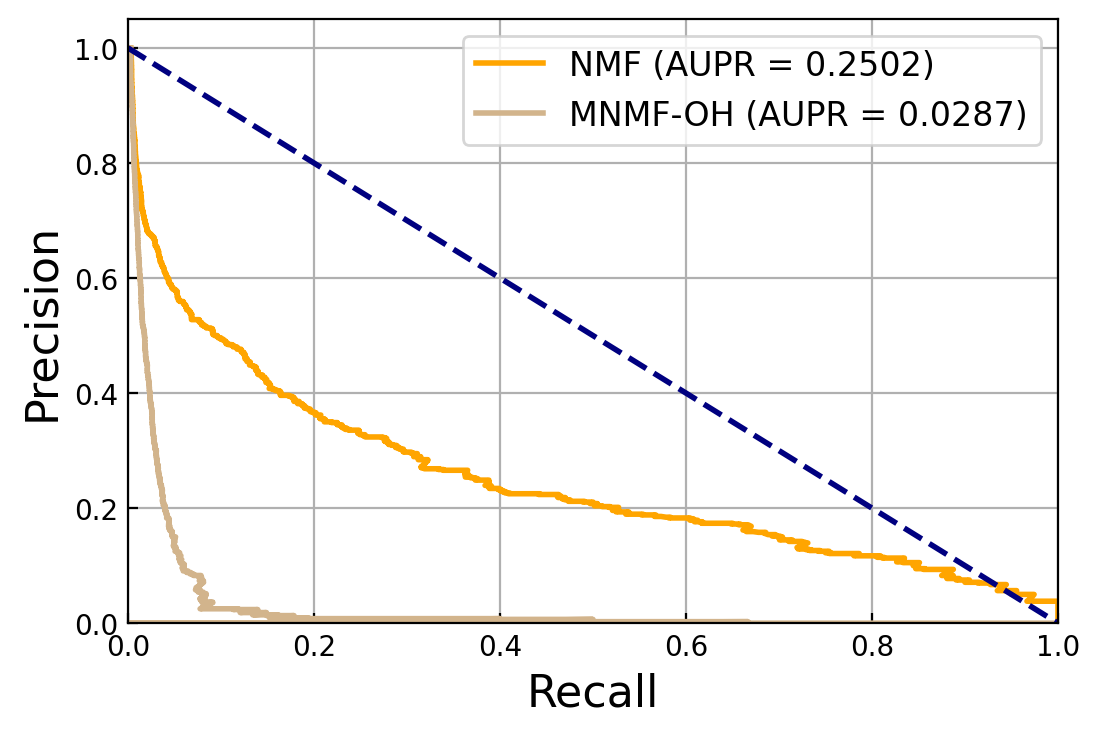}
	}
	\centering
	\caption{ The experimental comparison of neural metric factorization model with or without metrics information.}
\end{figure}\par

The purpose of embedding the two metrics information of drugs or diseases into the coding space of the latent factor vector is to make the points of similar drugs or diseases as close as possible. Let the latent factor vector be intuitively consistent with one of the assumptions of drug discovery, namely that "similar drugs treat similar diseases". \par

Therefore, in order to verify the effectiveness of the metric information, i.e., whether it can improve the prediction of the model, we run the following comparison experiments. The Neural Metric Factorization (One-Hot) refers to the NMF model that removes metric information. The results of the experiments are shown in Figure 8. We can find that the NMF model with embedded metric information outperforms the Neural Metric Factorization (One-Hot) model on both datasets, which indicates the effectiveness of our proposed improvement point. And the improvement of the NMF model is especially obvious under the AUPR metric compared to the AUC metric. This should be due to the embedding of metric information, which alleviates the sparse problem of computational drug repositioning dataset. The above two related experiments and analysis are the answer to \textbf{RQ2}.\par

\subsection{Comparison Method}

In order to fairly demonstrate the superiority of our proposed NMF model, we compare it experimentally with the following mainstream models. In addition, the latent facotor dimension of all models below is 32 and the number of negative samples is 5. \par

\begin{itemize}
	\item[\textbf{MF}] \cite{mf} The matrix factorization model is the most classical recommendation prediction model, which maps drugs and diseases into a low-dimensional vector space and represents the association between drugs and diseases by the inner product.
	\item[\textbf{NCF}] \cite{ncf} The neural collaborative filtering is a variant of the matrix factorization model, which optimizes the inner product operation using neural networks, allowing it to learn the nonlinear association between drugs and diseases.
	\item[\textbf{MetricF}] \cite{metricf} The metric factorization model uses Euclidean distances to express the interaction between the drug and the disease and factorizes it on the basis of a distance matrix to achieve a complementary matrix.
	\item[\textbf{ANMF}] \cite{m2} The additional neural matrix factorization model overcomes data sparsity using ancillary information on drugs or diseases and mitigates category imbalance using negative sampling techniques.
	\item[\textbf{NMF-OH}] The neural metric factorization (one-hot) is a variant version of the NMF model proposed in this paper, the main difference being that it does not use metric information to extract the latent factor of a drug or disease.
\end{itemize}

\begin{table}[]
	\caption{Comparison of experimental results}
	\centering
	\setlength{\tabcolsep}{2mm}
	\begin{tabular}{@{}cccccccc@{}}
		\toprule
		Dataset                                        & Evaluation Metrics & MF    & NCF   & MetricF & ANMF    & NMF-OH & NMF   \\ \midrule
		\multicolumn{1}{c|}{\multirow{2}{*}{Gottlieb}} & AUC                & 0.666 & 0.767 & 0.804   & 0.85   & 0.746  & 0.854\\
		\multicolumn{1}{c|}{}                          & AUPR               & 0.019 & 0.01  & 0.153   & 0.088  & 0.018  & 0.157 \\ \midrule
		\multicolumn{1}{c|}{\multirow{2}{*}{Cdataset}} & AUC                & 0.695 & 0.837 & 0.839   & 0.89  & 0.808  & 0.883 \\
		\multicolumn{1}{c|}{}                          & AUPR               & 0.021 & 0.023 & 0.232   & 0.161  & 0.028  & 0.25 \\ \bottomrule
	\end{tabular}
\end{table}

Table 2 shows the specific experimental results for each comparison model on the two datasets. First, our proposed NMF model achieved the best prediction results on all datasets. The AUC value on the Gottlieb dataset is 0.854 and the AUPR value is 0.157. The AUC value on the Cdataset is 0.883 and the AUPR value is 0.25. Second, although the prediction results of NMF and the SOTA model, ANMF, are similar under the AUC metric. However, considering that the AUC metric does not evaluate the data set of category imbalance well, more attention needs to be paid to the AUPR metric. Compared with the current SOTA model ANMF, the AUPR of the NMF model on the two data sets has increased by 66.8\% on average. This illustrates that our proposed NMF model has the better predictive capability and can better adapt to unknown data. Then, comparing the matrix factorization models like MF and NCF, the prediction ability of the NMF model is substantially ahead on both data sets, which indicates that the expression form of using points and generalized Euclidean distances in a high-dimensional coordinate system is more advantageous than the expression form like vector and inner product, which also verifies the effectiveness of the improvement points in this work. And it is worth noting that both the NMF model and the metric factorization model use distance to express the association between drugs and diseases, but the former performs significantly better than the latter. The reasons behind this are that there is an extreme class imbalance in the computational drug repositioning dataset, the metric factorization model destroys the original distribution of the data when converting the similarity matrix into a distance matrix, and the Euclidean distance is considered designed and does not have the ability to learn from the data. On the contrary, our proposed NMF model does not change the distribution of the data and the generalized Euclidean distance can automatically learn the distance from the data that matches its distribution, which consequently leads to the superior performance of the NMF model. The above related experiments and analysis are the answer to \textbf{RQ3}.\par


\section{Conclusion}

In this work, unlike the traditional matrix factorization model that uses vector and inner product to infer the association between drugs and diseases, we propose a novel neural metric factorization model, which represents drugs and diseases by points in a high-dimensional coordinate system and uses a novel generalized Euclidean distance to represent the association between drugs and diseases. And the metric information of drugs or diseases is used to control the range of variation of their positions in the high-dimensional coordinate system. Extensive experiments on two real data sets validate the effectiveness of the above two improvement points and verify the superiority of the NMF model by comparing the current mainstream prediction models. \par

However, the NMF model is essentially a variant belonging to the matrix factorization model and cannot effectively capture the local structural information of the strong drug-disease association. Therefore, in our future work, we will investigate how to embed models that can capture local structural information, such as random walk, into the NMF model. And how to better design the negative sampling strategy is also the starting point of our future work. \par




\bibliography{reference.bib}

\end{document}